\documentclass[letterpaper]{article} 
\usepackage[draft]{aaai25}  
\usepackage{times}  
\usepackage{helvet}  
\usepackage{courier}  
\usepackage[hyphens]{url}  
\usepackage{graphicx} 
\usepackage{natbib}  
\usepackage{caption} 
\usepackage{algorithm}
\usepackage{algorithmic}
\usepackage{booktabs}
\usepackage{tabularray}
\usepackage{xcolor}
\usepackage{subfig}
\usepackage[accsupp]{axessibility}
\usepackage{newfloat}
\usepackage{listings}
\DeclareCaptionStyle{ruled}{labelfont=normalfont,labelsep=colon,strut=off} 
\floatstyle{ruled}
\newfloat{listing}{tb}{lst}{}
\floatname{listing}{Listing}

\title{Try-On-Adapter: A Simple and Flexible Try-On Paradigm}
\author{
    {Hanzhong Guo\textsuperscript{\rm 2}\thanks{Equal contribution.}},
    Jianfeng Zhang\textsuperscript{\rm 2}{$^{*}$},
    Cheng Zou\textsuperscript{\rm 1}{$^{*}$}\thanks{Corresponding author.},
    Jun Li\textsuperscript{\rm 2},
    Meng Wang\textsuperscript{\rm 1},\\
    Ruxue Wen\textsuperscript{\rm 2},
    Pingzhong Tang\textsuperscript{\rm 2,\rm 3},
    Jingdong Chen\textsuperscript{\rm 1},
    Ming Yang\textsuperscript{\rm 1},
}

\affiliations {
    \ \\
    \textsuperscript{\rm 1} Ant Group \ \ 
    \textsuperscript{\rm 2} Turingsense \ \ 
    \textsuperscript{\rm 3} Tsinghua University\\
    \ \\
    guohanzhong@ruc.edu.cn, \  jianfengzhang2020@zju.edu.cn, \ 
 wuyou.zc@antgroup.com
}

\usepackage{bibentry}

\begin{document}

\maketitle

\begin{abstract}
Image-based virtual try-on, widely used in online shopping, aims to generate images of a naturally dressed person conditioned on certain garments, providing significant research and commercial potential. A key challenge of try-on is to generate realistic images of the model wearing the garments while preserving the details of the garments. Previous methods focus on masking certain parts of the original model's standing image, and then inpainting on masked areas to generate realistic images of the model wearing corresponding reference garments, which treat the try-on task as an inpainting task. However, such implements require the user to provide a complete, high-quality standing image, which is user-unfriendly in practical applications. In this paper, we propose Try-On-Adapter (TOA), an outpainting paradigm that differs from the existing inpainting paradigm. Our TOA can preserve the given face and garment, naturally imagine the rest parts of the image, and provide flexible control ability with various conditions, e.g., garment properties and human pose. In the experiments, TOA shows excellent performance on the virtual try-on task even given relatively low-quality face and garment images in qualitative comparisons. Additionally, TOA achieves the state-of-the-art performance of FID scores 5.56 and 7.23 for paired and unpaired on the VITON-HD dataset in quantitative comparisons.
\end{abstract}

\section{Introduction}
\label{sec:intro}

Image-based virtual try-on aims to naturally dress a model with given reference garment images, widely used in online shopping. 
A key challenge for try-on is to fit the non-rigid warping of a garment to a target body shape, while not making distortions in garment pattern and texture~\cite{han2018viton,choi2021viton,wang2018toward}.
Most existing try-on methods adopt multi-stage approaches~\cite{ge2021disentangled,han2018viton} which accomplish try-on tasks by including structure estimation, clothes warping~\cite {han2019clothflow,ge2021parser,wang2020linformer}, and image generation step by step, which couples the entire pipeline of try-on tasks into several subtasks. 


Recently, deep generative models (DGMs) and especially (score-based) diffusion models~\cite{sohl2015deep,ho2020denoising,song2020score,karras2022elucidating} have made remarkable progress in various domains, including text-to-image generation~\cite{ho2022cascaded, dhariwal2021diffusion,xu2022versatile,bao2023one}, audio generation~\cite{kong2020diffwave,popov2021grad}, video generation~\cite{ho2022video}, text-to-3D  generation~\cite{poole2022dreamfusion,wang2024prolificdreamer}. 
Beyond remarkable generation performance, another interesting aspect is that large-scale diffusion models provide rich prior knowledge. These rich priors lie in diffusion models that have already understood the color, the texture, and the pose. This eliminates the requirements to train the network from scratch to acquire a satisfactory performance. 
Previous studies complete the try-on tasks via stable diffusion models such as StableViton~\cite{kim2023stableviton} and Street TryOn~\cite{cui2023street}. 
These methods formulate the try-on tasks as inpainting tasks, which mask certain parts of the whole standing image and then generate the model's image wearing the reference garments utilizing the surplus unmasked areas.

However, such constraints of inpainting reduce the control ability in try-on tasks and are user-unfriendly due to the difficulty of obtaining high-quality standing images. 
Therefore, to achieve simplified input and flexible control for try-on tasks, we aim to find a paradigm that can refer to given items, such as the face and garment. Then, we seek to "imagine" the missing parts that are not provided. Finally, we desire to combine all these components to generate one realistic image. Such a paradigm is similar to outpainting, which involves extending the area outside a given image. The difference, however, is that we need to integrate a given reference image naturally.

To realize natural integration of different items in the outpainting paradigm and generate a realistic image, the image-as-prompt technique is a choice. Previous studies such as IP-Adapter~\cite{ye2023ip}, InstantID~\cite{wang2024instantid}, and PhotoMaker~\cite{li2023photomaker} are built upon the foundation of face recognition models by adding extra layers, enabling Stable Diffusion to interpret the images as text. As a result, joint image and text-guided generation can be achieved during the inference process.
Refer to these related works, we believe that the model can understand and combine items such as face images and garment images through some extra adapters and effective training, without destroying its rich priors and remarkable generation ability.

\begin{figure*}[tb]
    \centering
    \includegraphics[width=0.99\textwidth]{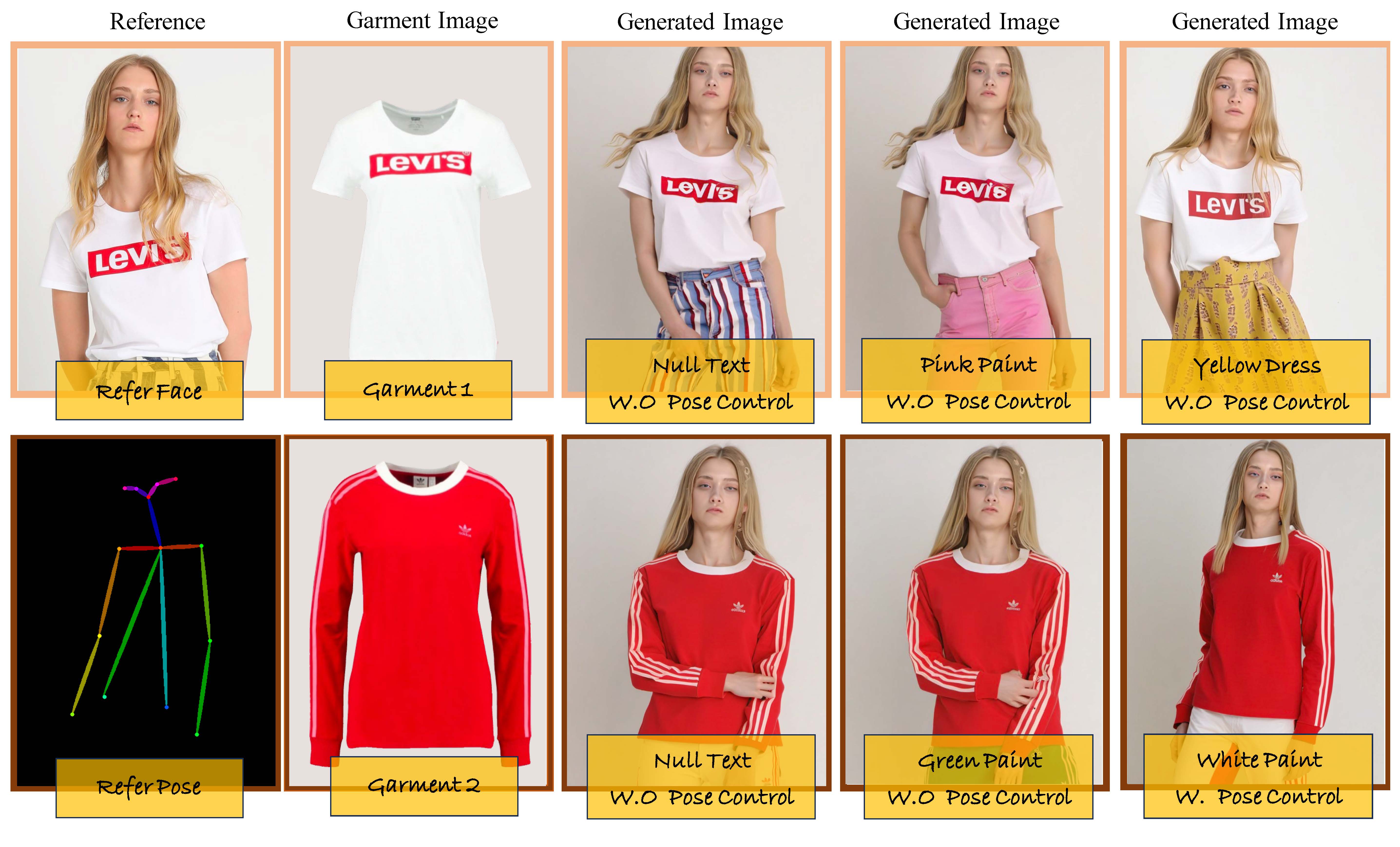}
    \caption{\textbf{Results of Try-On-Adapter.} 
    The first column gives reference face (top) and reference pose (bottom). The second column shows two target garments. 
    The third to fifth columns show try-on results generated by TOA with different conditions. The third column is generated by the reference face and target garments, conditioned on null text guidance, without pose control. The fourth column is generated by the reference face and target garments, conditioned on text guidance, without pose control. The last column is generated by the reference face and target garment, conditioned on both text guidance and pose control.}
    
    \label{fig:fig1}
\end{figure*}

In this paper, we reformulate the try-on task as an outpainting task, which generates realistic and naturally dressed images of the models via "imagine" the rest parts given the reference face and garment images, and propose the Try-On-Adapter (TOA) to accomplish the outpainting ability for try-on task.
To train the Try-On-Adapter, we put forward a pipeline of processing the original try-on dataset to obtain the facial image and corresponding prompt for the original image and then train the Try-On-Adapter via the denoising score matching with the processed dataset.

Compared with existing methods, Try-On-Adapter has the following threefold advantages: 
(1) Try-On-Adapter has a simpler input, it only requires a given facial image and a reference garment, instead of a high-quality standing image, which is much more user-friendly in practice. 
(2) Try-On-Adapter possesses a more flexible control ability, where users can edit the properties of garments and the human poses which is hard to change in traditional try-on methods, details in Fig.~\ref{fig:fig1}.
(3) Experimental results show that our Try-On-Adapter outperforms the latest OOTDiffusion \cite{xu2024ootdiffusion} in qualitative comparisons and achieves the state-of-the-art performance of FID scores 5.56 and 7.23 for paired and unpaired on the VITON-HD dataset.

\section{Related Work}

We focus on designing the try-on task as the outpainting task and proposing an image prompt adapter for handling and combining different reference images.
In this section, we
review recent works on the virtual try-on and text-to-image diffusion models, and relevant studies on image-guide generation.

\subsection{Image-Based Virtual Try-On}

Image-based virtual try-ons can be regarded as a conditional generation task that given the reference garment image $I_g$ and person image $I_p$ as the input, aimed to generate the try-on image $I_f = G(I_p,I_g)$, where $G$ denotes the generator or generation method using a pre-trained model.

To accomplish the try-on task, existing methods firstly provide a prior for guiding the deformation of garment, such as utilizing a combination of human body representation~\cite{gong2018instance,zhang2019progressively,guler2018densepose,ma2017pose} as input and predicting the spatial structural information of the human body.
Then, they warp the garment, which transforms the garment image to the spatial distribution under the try-on state. Given the garment images and human body features such as cloth-agnostic person representation obtained in the first step, some research~\cite{duchon1977splines,jaderberg2015spatial,li2019dense} propose to use a network to transform the spatial position of pixel points to the warped garment images. Finally, they mask the area where the original garment is and add the transformed garment into the corresponding area. 

A key open problem for virtual try-on is the need for non-rigid warping of the garment to fit the target body shape without introducing distortions in the garment pattern and texture. TryOnGAN~\cite{lewis2021tryongan} has demonstrated remarkably impressive results in the try-on task. However, it can be observed that in most of the showcased examples generated via TryOnGAN, the pose variation between the source image and target image is not significantly large. Additionally, the loss of details is a common occurrence. 

Meanwhile, TryOnDiffusion~\cite{zhu2023tryondiffusion} proposes a try-on method based on cascaded diffusion models, with the core idea being to obtain person image $I_a$ by taking a given person-specific image $I_p$ and removing the original garment and to obtain the final segmented garment image $I_g'$ based on the given garment image $I_g$. At the same time, the 2D pose keypoints $(J_p, J_g)$ of the person and garment images are predicted using existing methods, resulting in the input condition $C_{tryon}=(I_a, I_g', J_p, J_g)$. TryOnDiffusion then employs a cascaded diffusion model composed of a base diffusion model and two super-resolution diffusion models to generate the final try-on image $I_{tr}$. Each diffusion model can be represented as $\epsilon_t = \epsilon_{\theta}(z_t,t,C_{tryon},t_{na})$, where $t_{na}$ is the set of noise augmentation levels for different conditional images, and $z_t$ is the noisy images. The demonstrated results of their approach indicate that it can handle large occlusions, pose variations, and body shape changes while maintaining garment details at a resolution of 1024×1024.

With the development of large-scale generative models, it is possible to improve the performance of the try-on task due to the rich prior knowledge and controllability of large-scale diffusion models.
Street TryOn~\cite{cui2023street} utilizes the inpainting ability of Stable Diffusion to accomplish the try-on task.
For the initial garment image $I_g$ and its corresponding dense pose image $P_G$ and the target dense pose image $P_H$ obtained from the person-specific image $I_p$ are utilized to predict the warped target parse $M_T$. Subsequently, a small network refines the warped garment to obtain the refined warped garment $\omega(I_g,P_H)$. In another branch, the original garment mask of $I_p$ is removed, and the refined warped garment $\omega(I_g,P_H)$ is composited to form the composited person and garment image $I_T$. Finally, Stable Diffusion in combination with Controlnet~\cite{zhang2023adding} is employed to refine the image and obtain the final result.

\subsection{Diffusion Models}
Diffusion models gradually perturb data with a forward diffusion process and then learn to reverse such process to recover the data distribution. 
Formally, let $x_0\in\mathbb{R}^n$ be the unknown data distribution $q(x_0)$. 
Diffusion models define a forward process $\left \{ x_t \right \} _{t\in [0,1]}$ indexed by time $t$, which perturbs the data by adding noise to $x_0$ with the following equation, 
\begin{eqnarray}
\label{eq:dpmf}
\label{eq:forward}
     q(x_t|x_0) = \mathcal{N}(x_t|a(t)x_0,\sigma^2(t)I),
\end{eqnarray}
where $a(t), \sigma^2(t)$ denote the noise schedule which controls the noise level added to the data, and $a(t)$ gradually decreases to zero to transform the data into a standard Gaussian distribution.
Meanwhile, to reverse the forward process in Eq.~\eqref{eq:dpmf}, DDPM~\cite{ho2020denoising} puts forward the Eq.~\eqref{eq:dt_reverse} which defines the reverse transition kernel as a Gaussian distribution as the following,

\begin{align}
\label{eq:dt_reverse}
p(x_{t-1}|x_t) = \mathcal{N}(x_{t-1}|\frac{1}{\alpha(t)} & \nonumber (x_t-\frac{1-\alpha^2(t)}{\sqrt{1-\bar{\alpha}^2(t)}} \\ &\epsilon_{\theta}(x_t,t)),\Sigma_{t}I) ,
\end{align}
where $\Sigma_{t}$ is the hand-crafted variance in the reverse transition kernel such as taken as the $\sigma^2(t)$, ScoreSDE~\cite{song2020score} proved that given one noise schedule, the forward process in Eq.~\eqref{eq:forward} equals to one specific stochastic differential equation (SDE) and the reverse process in Eq.~\eqref{eq:dt_reverse} is the Euler discretization of the corresponding reverse SDE. To sample via Eq.~\eqref{eq:dt_reverse}, the only unknown is the noise network $\epsilon_{\theta}(x_t,t)$, which can be learned by denoising score matching (DSM) loss in DDPM~\cite{ho2020denoising},
\begin{eqnarray}
\label{eq:loss}
    &\mathcal{L}(\theta) = \int_{0}^1w(t)\mathbb{E}_{q(x_0)}\mathbb{E}_{q(\epsilon )}[\left \| \epsilon _{\theta}(x_t,t)-\epsilon  \right \|_2^2 ]d t ,
\end{eqnarray}
where $w(t)$ is a weighting function, $q(\epsilon)$ is standard Gaussian distribution and $x_t \sim q(x_t|x_0)$ follows Eq.~\eqref{eq:forward}. 

Besides unconditional sampling in Eq.~\eqref{eq:dt_reverse}, another important application for diffusion models is conditional sampling, especially the text-to-image diffusion models.
The text-to-image task aims to synthesize realistic and high-resolution images from text prompts $y$. 
Among the works in the text-to-image generation, most of the existing methods are based on latent diffusion models (LDMs)~\cite{rombach2022high, podell2023sdxl}, 
which diffuses in the latent space of one variational autoencoder (VAE), and the noise network in latent diffusion models conditioned on the noisy image, timestep, and the corresponding text prompt as follows,
\begin{align}
\label{eq:sd_loss}
    \mathcal{L}(\theta)_{\textrm{LDM}} = &\int_{0}^1w(t)\mathbb{E}_{z_0,y \sim [\varepsilon (q_0),y]}\mathbb{E}_{q(\epsilon )}\\ \nonumber &[ \left \| \epsilon _{\theta}(z_t,t,y)-\epsilon  \right \|_2^2 ]d t ,
\end{align}

where $\varepsilon$ denotes the VAE, and $z_t=\alpha(t)z_0+\sigma(t)\epsilon$ which diffuses follows Eq.~\eqref{eq:forward}, $y$ is the text prompt, $q_0$ denotes the data distribution. After training, $z_0$ generated via Eq.~\eqref{eq:dt_reverse} and decode to the original data space $x_0 = D(z_0)$.



\subsection{Image-guided Generation}
Despite the considerable success of text-to-image diffusion models, relying solely on text guidance to generate high-quality images requires complex and specialized prompt engineering to fully unlock their potential. 
An alternative to text-guided generation is image-guided generation~\cite{wang2024instantid, hu2024instruct, li2024blip, pan2024generating,ye2023ip}. The core idea behind the image-guided generation is to embed the image into the text embedding which means that taking the image as a prompt and sample via the corresponding noise network $\epsilon _{\theta}(z_t,t,y,\mathbf{E}(I))$, where $y$ is the text prompt and $I$ denotes the reference image, $\mathbf{E}$ denotes the image embedding above.

To be specific, IP-Adapter~\cite{ye2023ip} proposes an effective and lightweight adapter to achieve image-guided capability and leverages a novel decoupled cross-attention mechanism to embed image features through additional cross-attention layers to achieve image prompts in parallel with text prompts. 
InstantID~\cite{wang2024instantid} designs a novel IdentityNet based on IP-Adapter, which integrates reference images with text prompts to guide image generation by imposing strong semantic and weak spatial conditions. 



\section{Method}
While most existing methods for virtual try-on use generative adversarial networks or diffusion models via inpainting, we propose Try-On-Adapter, which to the best of our knowledge is the first solution based on the outpainting paradigm.
To implement Try-On-Adapter, we first process original try-on dataset to obtain facial images and text prompts via pre-trained models. Then, we use the garment images, facial images and text prompts to train Try-On-Adapter with denoising score matching loss. 


\begin{figure*}[tp]
    \centering
    \includegraphics[width=0.99\textwidth]{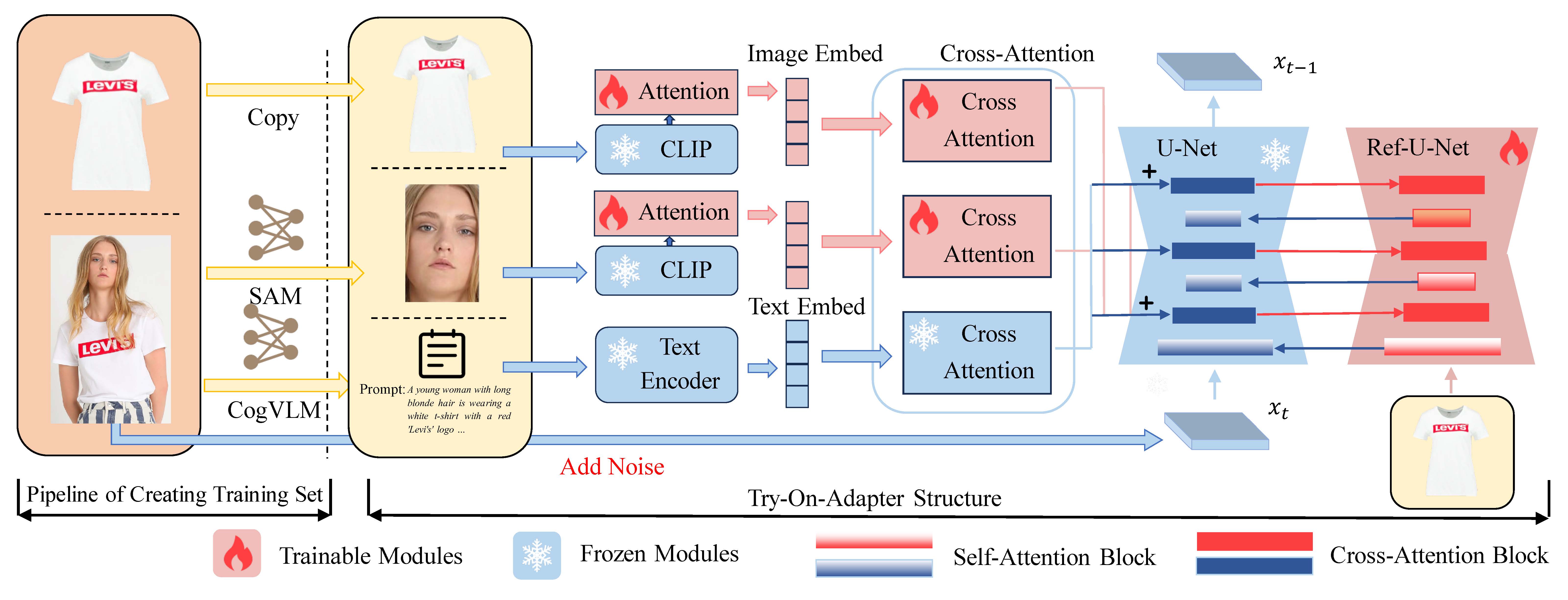}
    \caption{\textbf{Overall architecture of Try-On-Adapter.} 
    The Blocks in red represent the trainable blocks while the blocks in blue denote the frozen pretrained blocks. 
    The blocks in orange represent the original available public try-on dataset and the blocks in yellow represent the obtained dataset.
    The proposed Try-On-Adapter primarily comprises two modules: the fusion of different references on the Cross-Attention Block and the Reference U-Net specifically designed for garments.
    }
    \label{fig:architecture}
\end{figure*}

\subsection{The Architecture of Try-On-Adapter}
\label{sec:loss}
Since existing inpainting paradigm solving try-on tasks are user-unfriendly, requiring high-quality standing images, and reducing the flexible control ability, we desire to design a try-on algorithm based on the outpainting paradigm. 
To implement the try-on using the concept of outpainting, we need to address two issues. The first is how to efficiently extract the representations of different items using pre-trained models, and the second is how to naturally integrate these representations into the diffusion model to achieve the desired try-on effect. 
To address these two issues, we have made improvements to both the cross-attention block and the self-attention block of the original U-Net.

\subsubsection{Comprehension- and Fusion-inspired Cross-attention.}
IP-Adapter~\cite{ye2023ip} and InstantID~\cite{wang2024instantid} have proved that an extra small module based on a pre-trained vision model can handle complex information in reference images. Both of these researches involve encoding the image into the latent space, and then inserting it as a text embedding into the cross-attention block of the U-Net. 
Their experiments have demonstrated that inserting image information into the cross-attention block can effectively maintain image features while allowing for natural combinations.
In our paper, we refer to the IP-Adapter that uses the CLIP~\cite{radford2021learning} as the pre-trained vision model since CLIP can handle both color information and texture information, which is essential in the try-on setting. 
Due to the gap between the text embedding in pre-trained text-to-image models and the latent space of images after CLIP encoding, IP-Adapter employs an extra module that projects the image latent space in CLIP onto the text embedding in the text-to-image diffusion models, which acts for the goal of Image-As-Prompt. In IP-Adapter, there are two kinds of modules, one is to encode the image to obtain the embedding of \textbf{[CLS]} token and learn a linear projection, and another is to use the whole hidden state $\Pi \in \mathbb{R}^{[\textrm{N},\textrm{M}]}$ of image in CLIP embedding where N denotes the number of tokens and learn a small transformer block to handle the information in the whole hidden state. 
The experiments of IP-Adapter have shown that the second approach can better preserve the details of the images. In the try-on task, the changes in garment details are greater than the facial information in the IP-Adapter. Therefore, we adopted the latter scheme.

To be specific, we use a small attention block to fusion the whole hidden state and project the output of the attention block into the text embedding $\varepsilon \in \mathbb{R}^{[\textrm{n},\textrm{m}]}$ via a linear projection, where n denotes the number of tokens and m denotes the dimension of latent text embedding. The architecture of the proposed attention block is shown in Eq.~\eqref{resample},
\begin{align}
\label{resample}
\Pi_{l+1} = &\textrm{Atten}(\Pi_{l}(I_g,I_f)) \\ \nonumber = &\textrm{Atten}(\textrm{Q}_{in}(L),\textrm{K}_{in}(\Pi_{l},L),\textrm{V}_{in}(\Pi_{l},L)),
\end{align}
where $\Pi_l$ represents the $l$-th layer output of the proposed attention block conditioned on the whole hidden state $\Pi(I_g,I_f)$ of the CLIP, $I_g$ and $I_f$ represent the garment image and face image, respectively.
$L$ is the learnable parameters with dimension $\mathbb{R}^{[n,m]}$.
And $\textrm{Q}_{in}=LW_q, \textrm{K}_{in}=[\Pi_n,L]W_k, \textrm{V}_{in}=[\Pi_n,L]W_v$, where $W_q, W_k, W_v$ are the linear projection $\mathbb{R}^{[h,l]} \longmapsto \mathbb{R}^{[h,l]}$.
$\textrm{Atten}$ is the attention block~\cite{vaswani2017attention}. 
After attention calculation in Eq.~\eqref{resample}, 
the output will be reshaped to dimension $\mathbb{R}^{[n,k]}$, where $n$ denotes the number of tokens representing the image. 
Finally, to match the dimension of text embedding, we add a feedforward $\mathbb{f}: \mathbb{R}^{[n,k]} \longmapsto \mathbb{R}^{[n,m]}$ to transform. Therefore, to obtain the latent embedding of reference images, we learn the parameters in $L$, $W_q, W_k, W_v$, $\mathbb{f}$ and meanwhile keep other parameters frozen. 
The proposed attention block in the Try-On-Adapter is located as shown by the red cross-attention block in Fig.~\ref{fig:architecture}.

Another important issue is how to naturally integrate the representation of different items. 
In our paper, we follow the IP-Adapter to use the decoupled cross-attention~\cite{ye2023ip}, expressed in Eq.~\eqref{decouple}.
\begin{align}
\label{decouple}
Z=\Pi^{\textrm{text}}_l+\Pi^{\textrm{face}}_l+\Pi^{\textrm{garment}}_l,
\end{align}
where $l$ denotes the depth of attention block in Eq.~\eqref{resample}, $\Pi^{\textrm{face}}_l$, $\Pi^{\textrm{garment}}_l$ denote the embedding calculated from Eq.~\eqref{resample} conditioned on $I_f$ and $I_g$, respectively. 
$Z$ is the cross-attention block output in the Try-On-Adapter, which will be fed into the subsequent layers of the U-Net.

\subsubsection{Reference-preserved Self-attention.} 
In the experiments, we found that while the facial information is well preserved when fusing image information in the aforementioned cross-attention block, some details of the garment tend to be lost. 
This is partly related to the reference image being resized to a 224-pixel resolution when entering CLIP, during which some complex garment details may be lost. 
Furthermore, we discovered that when optimization is performed solely in the cross-attention block, the garment patterns often change. 
More details can be found in Appendix. A. Therefore, to better encode the image information of the garment, we referred to the approach of AnimateAnyone~\cite{hu2023animate} and introduced the Reference U-Net.

Following AnimateAnyone~\cite{hu2023animate}, we adopt a framework identical to the denoising UNet for Reference U-Net and inherit its weights from the original Stable Diffusion.
Then, in each self-attention block of the original U-Net, the input will be concatenated with the self-attention block of the corresponding layer from the Reference U-Net.
To be specific, given a feature map $u_1 \in \mathbb{R}^{h \times w \times c}$ from denoising UNet and $u_2 \in \mathbb{R}^{h \times w \times c}$ from Reference U-Net, we concatenate it with $u_1$ along $w$ dimension. 
Then we perform self-attention and extract the first half of the feature map as the output. 
Additionally, cross-attention is employed using the output of our proposed cross-attention block which conditioned the text, garment reference image, and face image. 


To train the parameters mentioned above, we adopt the following objective function, 
\begin{small}
\begin{eqnarray}
\label{eq:toa_loss}
    \mathcal{L}(\theta) = &\int_{0}^1w(t)\mathbb{E}_{z_0,y,I_f,I_g \sim [\varepsilon (q_0),y,I_f,I_g]} \mathbb{E}_{q(\epsilon )}\\ \nonumber &[\left \| \epsilon _{\theta}(z_t,t,y,\textrm{TOA}(I_f,I_g))-\epsilon  \right \|_2^2 ]d t ,
\end{eqnarray}
\end{small}
where $\textrm{TOA}$ represents the proposed Try-On-Adapter, and the combination of text prompt $y$ and $\textrm{TOA}(I_f,I_g)$ is the decoupled cross-attention in Eq.~\eqref{decouple}. To utilize the classifier-free guidance in sampling, we randomly drop the image embedding and replace it with zero embedding.



\subsection{Construct the Training Data}
\label{sec:train}
Try-On-Adapter requires a reference facial image instead of a high-quality whole standing image for training, resulting in a data requirement discrepancy with the current public datasets, e.g., VITON-HD~\cite{choi2021viton}.
As shown in Fig.~\ref{fig:architecture}, Try-On-Adapter supports reference facial image and garment image as conditions, while none of the existing datasets contains facial boxes, so we use SAM~\cite{kirillov2023segment} to obtain the facial area to get reference facial images. Further, we randomly expand the facial area to increase the diversity of the reference facial images. This random facial image augmentation dramatically improves the generalization of the model. If it expands slightly, the model will be robust to the reference facial images, but if it expands a lot, even to the whole image, it can support a whole standing image as a reference. From this point of view, traditional methods those conditioned on the whole standing image are just a special case of our method.

Try-On-Adapter is also designed to support precise text prompt guidance. Although employing null text condition has already enabled the try-on ability, we need more editability. ~\cite{BetkerImprovingIG} proved that improving the caption quality can improve the prompt following ability, thus we utilize CogVLM~\cite{wang2023cogvlm} to assign caption for each image to get better editability via text prompt.
As an example, we engage in a round of dialogue with CogVLM, asking, "Please provide a detailed description of this image, including the hairstyle, hair color, and the colors of tops and pants.", and then it outputs a detailed caption of the given image.

After undergoing the aforementioned processes, each image from the original dataset is transformed to a training sample for Try-On-Adapter, $C_{TOA}(I_g,I_f,p)$, where $I_g$ represents the garment image, $I_f$ represents the reference facial image obtained via the SAM, $p$ is the text prompt from CogVLM. The left part of Fig.~\ref{fig:architecture} illustrates the pipeline of creating training set.

\subsection{Training and Inference}
\label{train and inf}
Due to the high cost of obtaining high-quality try-on data, we design a two-stage training schedule to improve the generalization ability of the Try-On-Adapter. The first stage utilizes easily available Internet data for pretraining, and the second stage utilizes a small amount of high-quality data for finetuning. Both the Internet data and the high-quality data are processed in the same way as mentioned in the previous section.

Technically, in the first stage, we collect 0.2M full-body and half-body images from the Internet and use SAM to segment the face and garment area. 
Due to the varying quality of the data, there inevitably exist 
segmentation failures, resulting in a relatively low data quality. 
Therefore, we use this data for the first-stage pretraining, mainly to enable the model to understand faces and garments and build the combination ability.
After completing the first training stage, we finetune the model on the high-quality try-on dataset, e.g., VITON-HD, which contains high-quality reference garment images and whole standing images. After two-stage training, our Try-On-Adapter can achieve the try-on functionality while performing generalization and editability, more details are shown in the experimental parts.

In the inference stage, the reference facial image and the garment image can be obtained from the Internet (e.g., screenshot), which are easier to acquire. With the null text prompt, the model can output a basic result of try-on. With additional text prompts as guidance, the properties, such as the color and style of the garment, can be easily edited.
Additionally, Try-On-Adapter is compatible to ControlNet so that more control conditions, such as human poses, can be used in try-on tasks.




\section{Experiments}

\subsection{Experimental settings}
Datasets and Implementation Details are shown in the Appendix. 

\subsection{Qualitative Results}
\label{sec:qualitative}
We qualitatively evaluate the Try-On-Adapter in two settings. One is the single dataset evaluation where the facial image and the garment image are from a single dataset. The other setting is the cross dataset evaluation where the two items are obtained from different datasets. 
To better evaluate the generalization and practicality of the Try-On-Adapter, the garment images used in the cross dataset evaluation are selected from the Internet, e.g., Civitai~\cite{civitai}, and captured as screenshots.
Additionally, comparisons are conducted with the most recently open-sourced OOTDiffusion~\cite{xu2024ootdiffusion}.

\begin{figure*}[t]
    \centering
    \subfloat[Single Dataset Evaluation]{
        \includegraphics[width=0.46\linewidth]{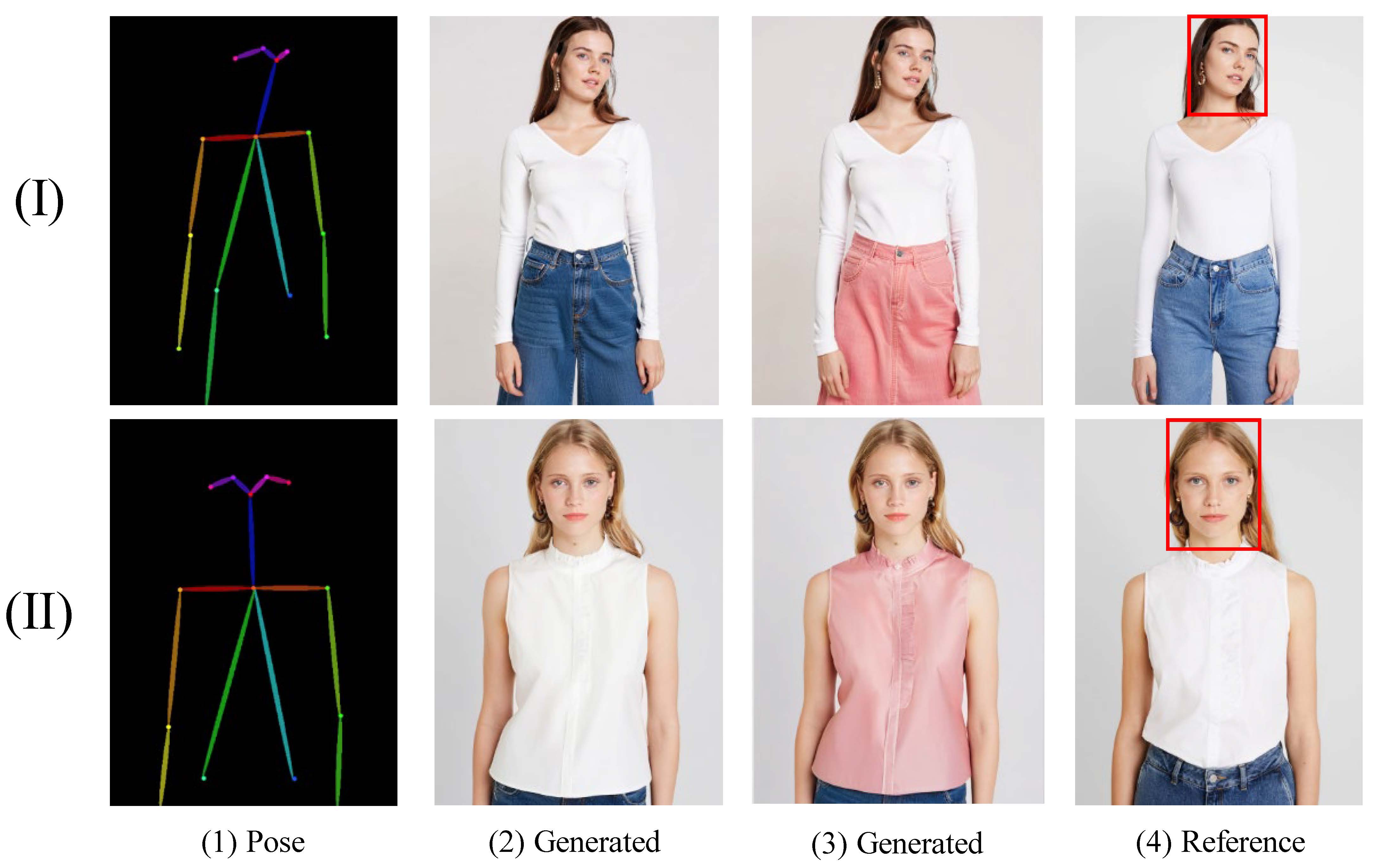}
        }
    \subfloat[Cross Dataset Evaluation]{
        \includegraphics[width=0.495\linewidth]{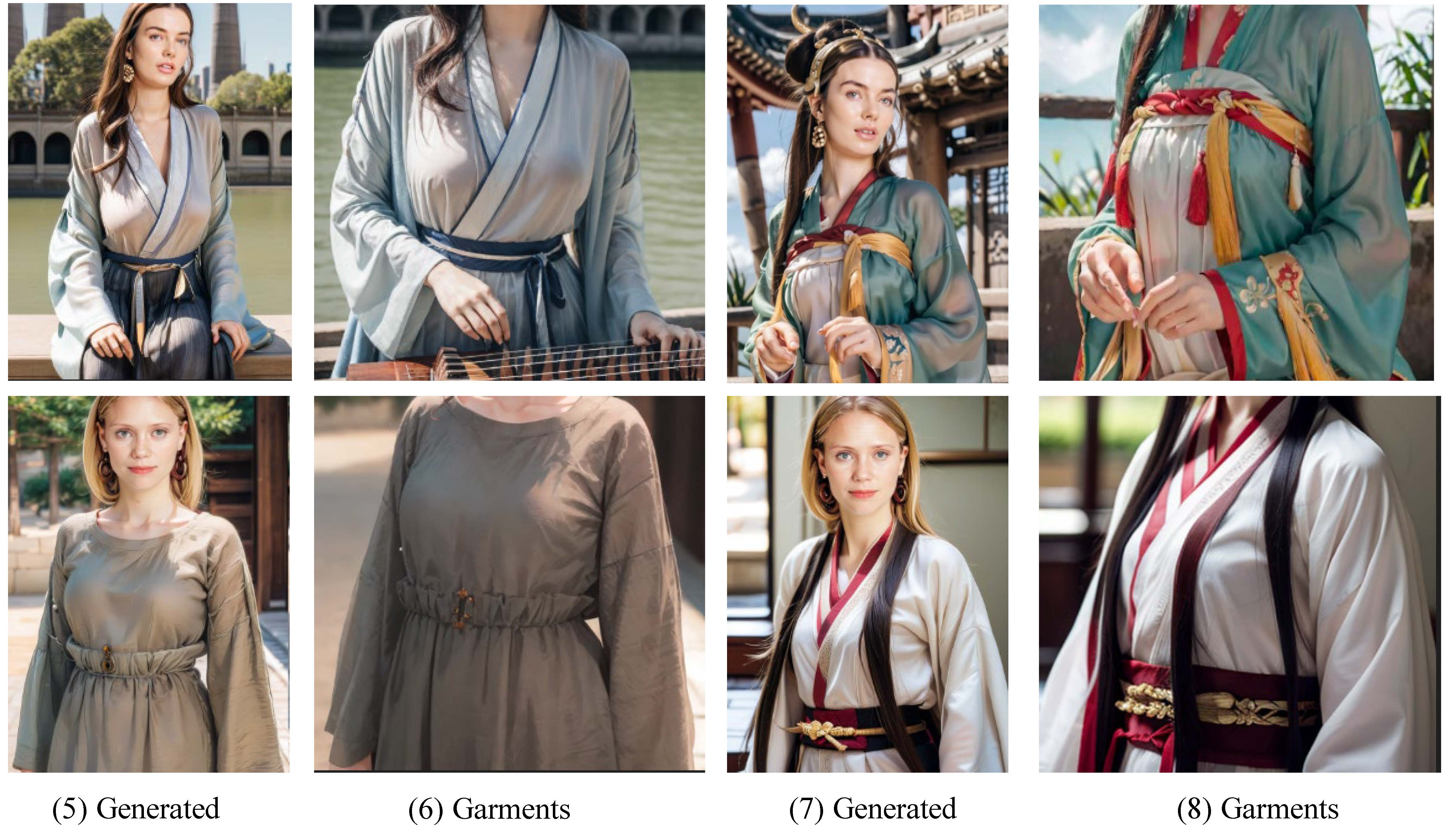}
    }
    \captionsetup{font={small}}
    \caption{
    \textbf{Qualitative Results of the Try-On-Adapter.} (a) is the single dataset evaluation where the faces and the garments are from the VITON-HD and (b) denotes the cross dataset evaluation where the faces are from the VITON-HD and the garments are from the Internet. (a)-(2) denotes the generation of TOA while given the pose in (a)-(1) and the face in (a)-(4). (I)-(a)-(3) denotes the generation under text guidance "pink dress"  and (II)-(a)-(3) under "pink tops". The generations in (b)-(5), (b)-(7) are conditioned on the garments from (b)-(6), (b)-(8), respectively. }
    \label{fig:qualitative}
\end{figure*}

\noindent\textbf{Single Dataset Evaluation.} Fig.~\ref{fig:qualitative}(a) shows the performance of Try-On-Adapter in the single dataset evaluation in which the facial image and the garment image both come from VITION-HD.
We set the reference facial image and the garment image as input, and then utilize the reference pose image to guide the generation via ControlNet in this setting.
Based on these three kinds of inputs, Fig.~\ref{fig:qualitative}(a)-(2) are generated via a null text prompt, it can be observed that the generated images are realistic while preserving the characters of the reference facial image and the reference garment which are shown in Fig.~\ref{fig:qualitative}(a)-(4), 
indicating that the Try-On-Adapter has a good capability of item preservation and combination.
Furthermore, to verify the control ability of the Try-On-Adapter, we add text prompt guidance, Fig.~\ref{fig:qualitative}(I)-(a)-(3) shows an example of "pink dress" and "pink tops" in Fig.~\ref{fig:qualitative}(\textrm{II})-(a)-(3).
It can be seen that the Try-On-Adapter responds well to the text guidance while maintaining its original try-on capability.

\noindent\textbf{Cross Dataset Evaluation.} In the cross dataset evaluation, 
the inputs are also facial image and the garment image, but the garment image comes from the Internet. Fig.~\ref{fig:qualitative}(b) shows the performance of the Try-On-Adapter. 
To better align with real-world applications, we capture garment images from the Internet, shown in Fig.~\ref{fig:qualitative}(b)-(6) and Fig.~\ref{fig:qualitative}(b)-(8). The reference facial images are cropped from the VITON-HD dataset. Results in Fig.~\ref{fig:qualitative}(b)-(5) and Fig.~\ref{fig:qualitative}(b)-(7) demonstrate that given street scene garments, the Try-On-Adapter can generate high-fidelity images while preserving the details of the garments, even if there exists domain gap between the item images, i.e., the reference facial images from VITON-HD and the garment images from the Internet.

\noindent\textbf{Qualitative comparisons.} 
To better evaluate the performance of Try-On-Adapter, we conduct qualitative comparisons with the state-of-the-art open-source project OOTDiffusion~\cite{xu2024ootdiffusion}, which are shown in Fig. 4 (in supplementary materials).
In Fig. 4, the Try-On-Adapter is conditioned on the second column (garment) and the third column (reference facial image), while the OOTDiffusion is conditioned on the first column (reference whole standing image) and the second column. 
From the results, it can be seen that when using the garments from VITON-HD as guidance (top two rows in Fig. 4), the Try-On-Adapter can generate results that are more consistent with the garments and without any hallucinations. 
Moreover, as shown in the third row in Fig. 4, when the reference standing image of OOTDiffusion is changed to a facial image, it fails to produce a normal output image. However, as shown in the fourth row, when the reference facial image of Try-On-Adapter changes to a standing image, it can still generate realistic and consistent results.
In the last row in Fig. 4, we evaluate the performance when the garment images are obtained from the Internet, and the Try-On-Adapter still shows better results.


\subsection{Quantitative Results}
Besides the qualitative results, we also evaluate the Try-On-Adapter with some widely used metrics.
Specifically, SSIM~\cite{wang2004image} and LPIPS~\cite{zhang2018unreasonable} are used to evaluate the similarity between the generated images and the ground truth, FID~\cite{heusel2017gans} is used to evaluate the realism of the generated images.
In this section, quantitative results are evaluated on the VITON-HD 6K test set, and the experiments are conducted under two settings, paired evaluation and unpaired evaluation. 

\label{sec:quantitative}

\begin{table*}[htbp]
\centering
\caption{\textbf{Quantitative comparisons on the VITON-HD dataset.} From the results, the Try-On-Adapter can generate the most realistic results among different methods in both paired evaluation and unpaired evaluation. The lower similarity with the original image is mainly due to the difference between the inpainting and outpainting paradigms, which has been discussed in Quantitative Results. }
\begin{tblr}{
  cell{1}{3} = {c=3}{},
  cell{1}{7} = {c=3}{},
  cell{3}{1} = {r=4}{},
  vline{2} = {1-7}{},
  hline{1,3,7-8} = {-}{},
  hline{2} = {1-5,7-9}{},
}
 \centering{Datasets}&  & Paired  &  &  &  & Unpaired &  &  \\
 \centering{Paradigms}& \centering{Method}  &\centering{SSIM$\uparrow$}  &\centering{LPIPS$\downarrow$}  & \centering{FID$\downarrow$} & &  &\centering{FID$\downarrow$}  \\
 \centering{Inpaint}&\centering{VITON-HD~\cite{choi2021viton}}  & 0.862 & 0.117 &  12.11 &  &  & 44.25  \\
 &\centering{StableVITON~\cite{kim2023stableviton}} & 0.852 & 0.084 & 8.698 & & & 12.58 \\
 &\centering{TryOnDiffusion~\cite{zhu2023tryondiffusion}} & - & - & 13.44 & & & 23.35 &   \\
 &\centering{OOTDiffusion~\cite{xu2024ootdiffusion}} &  \textbf{0.877} & \textbf{0.071} & 8.81 & & &  11.96 &   \\
 \centering{Outpaint}&\centering{TOA ~\textbf{(Ours)}}  & 0.772 & 0.178 & \textbf{5.56} & &  & \textbf{7.23} &  \\
\end{tblr}
\label{tab:results}
\end{table*}

\noindent\textbf{Paired Evaluation.} 
For paired evaluation, we generate the try-on image conditioned on the garment image and its corresponding facial image.
In this setting, we can evaluate the similarity between the generated image and the original image (ground truth), also we can evaluate the realism of the generated images by calculating the FID score between the set of generated images and the set of original images. As shown in Tab.~\ref{tab:results}, in the paired setting, TOA gets the smallest FID of 5.56, implying it generates the most realistic image among different try-on methods. 
However, the TOA generates less similar images compared to ground truth, which is mainly due to the difference of paradigms. Specifically, for the inpainting task most of the image regions are maintained (unchanged), so the similarity with the original image is higher, while for the outpainting task, most of the image regions are generated (changed), so the similarity with the original image is lower.
Since our TOA follows the outpainting paradigm, the pose, face, and other missing parts of the image can not be ensured to be consistent with the ground truth, leading to a lower similarity. On the other hand, this implies that TOA has the potential to provide more diversity.

\noindent\textbf{Unpaired Evaluation.} 
For unpaired evaluation, the reference facial image and the garment image come from different images, which is similar to real-world applications. In this setting, the reference facial image for TOA comes from one certain image, and the garment image is randomly selected from the test set to form a combination. In contrast, other methods choose a certain reference standing image and randomly select the garments from the test set. Since there is no ground truth image in this scenario, only the FID score can be evaluated. As can be seen from Tab.~\ref{tab:results}, TOA outperforms other methods on FID, implying that it generates the most realistic images, thus demonstrating its advantages in real-world applications.

\section{Conclusion}
In this paper, we propose Try-On-Adapter (TOA), a new paradigm based on outpainting, for virtual try-on. Compared with existing inpainting try-on methods, our TOA is more user-friendly and possesses a more flexible control ability.
Specifically, TOA can generate natural and realistic try-ons when only given the reference facial image and garment image. It also allows users to edit the properties (such as color and style) of the garments and the human poses.
The experimental results show that the proposed TOA generates more realistic try-ons than other approaches especially when the reference image is low-quality face or garment screenshots. Additionally, the TOA also achieves the SOTA performance of FID scores 5.56 and 7.23 for paired and unpaired evaluation on the VITON-HD dataset in quantitative comparisons.
We believe that with higher quality of data, the TOA can achieve finer-grained items combination generation, such as hats, tops, trousers, and shoes, which will be explored in the future.

\bibliography{aaai25}


\newpage
\onecolumn

\end{document}